\pdfoutput=1
\documentclass[11pt, a4paper]{article}
\usepackage{authblk}
\usepackage{acl}
\usepackage{times}
\usepackage{latexsym}
\newtheorem{definition}{Definition}
\usepackage[T1]{fontenc}
\usepackage[utf8]{inputenc}
\usepackage{microtype}
\usepackage{times}
\usepackage{latexsym}
\usepackage{url}
\usepackage{amsmath}
\usepackage[nameinlink]{cleveref}
\crefformat{section}{\S#2#1#3} % see manual of cleveref, section 8.2.1
\crefformat{subsection}{\S#2#1#3}
\crefformat{subsubsection}{\S#2#1#3}
\crefformat{paragraph}{\S#2#1#3}
\usepackage{enumitem}
\usepackage[colorinlistoftodos]{todonotes}
\usepackage{booktabs}
\usepackage{multirow}
\usepackage{fixltx2e}
\usepackage[most]{tcolorbox}
\usepackage{adjustbox}
\usepackage{booktabs}
\usepackage[font=small,labelfont=bf]{caption}
\usepackage{subcaption}
\usepackage{soul}
\usetikzlibrary{shapes}

\newcommand\GB[1]{\colorbox{green!30} {#1}}

\newcommand\RB[1]{\colorbox{red!30} {#1}}

\newcommand{\quoref}{\text{Quoref}}
\newcommand{\squad}{\text{SQuAD}}
\newcommand{\hotpot}{\text{HotpotQA}}
\newcommand{\wiki}{\text{2WikiMultiHopQA}}

\newcommand{\counterfactual}{\text{CF}}

\newcommand{\finetuned}{\text{fine-tuned}}
\newcommand{\finetune}{\text{fine-tune}}
\newcommand{\rl}{\text{RoBERTa\textsubscript{large}}}
\newcommand{\bertlc}{\text{BERT\textsubscript{large-cased}}}
\newcommand{\bertbc}{\text{BERT\textsubscript{base-cased}}}
\newcommand{\bertmedium}{\text{BERT\textsubscript{medium}}}
\newcommand{\bertsmall}{\text{BERT\textsubscript{small}}}
\newcommand{\roberta}{\text{RoBERTa}}
\newcommand{\bert}{\text{BERT}}

\newcommand{\fscore}{\text{F1-Score}}
\newcommand{\emscore}{\text{Exact-match}}
\newcommand{\ig}{\text{IG}}
\newcommand{\occlusion}{\text{Occlusion}}

\newenvironment{itemize*}%
  {\begin{itemize}%
    \setlength{\itemsep}{0.9pt}%
    \setlength{\parskip}{0.9pt}%
    \setlength{\topsep}{0.9pt}}%
  {\end{itemize}}

\usepackage{mdframed}
\AtBeginEnvironment{blockquote}{\small}
\definecolor{block-gray}{gray}{0.85}
\newtcolorbox{blockquote}{colback=block-gray,boxrule=0pt,boxsep=0pt,breakable}
\AtBeginEnvironment{definition}{\small}

\title{Machine Reading, Fast and Slow: \\ When Do Models ``Understand'' Language?}

\author[1*, 2]{Sagnik Ray Choudhury\Thanks{\enspace * Work done while employed at the University of Copenhagen}}
\author[2]{Anna Rogers}
\author[2]{Isabelle Augenstein}
\affil[1]{University of Michigan}
\affil[2]{University of Copenhagen}
\affil[ ]{\tt{sagnikrayc@gmail.com}, \tt{arogers@sodas.ku.dk}, \tt{augenstein@di.ku.dk}}

\date{}
\makeatletter
\def\thanks#1{\protected@xdef\@thanks{\@thanks
        \protect\footnotetext{#1}}}
\makeatother
\begin{document}
\maketitle

\begin{abstract}
Two of the most fundamental challenges in Natural Language Understanding (NLU) at present are: (a) how to establish whether deep learning-based models score highly on NLU benchmarks for the `right' reasons; and (b) to understand what those reasons would even be. We investigate the behavior of reading comprehension models with respect to two linguistic `skills': coreference resolution and comparison. We propose a definition for the reasoning steps expected from a system that would be `reading slowly', and compare that with the behavior of five models of the \bert\ family of various sizes, observed through saliency scores and counterfactual explanations. We find that for comparison (but not coreference) the systems based on larger encoders are more likely to rely on the `right' information, but even they struggle with generalization, suggesting that they still learn specific lexical patterns rather than the general principles of comparison.
\end{abstract}

\section{Introduction}
\label{sec:intro}

Generally, human decisions may be based on deliberate, careful reasoning (`slow thinking') or quick heuristics (`fast thinking') \cite{Kahneman_2013_Thinking_fast_and_slow}. These two processes have parallels in the realm of reading comprehension (RC): a human reader would ideally fully process the text to answer questions, but in practice, we may deliberately skim rather than read to save effort. Even capable students may be misled by superficial cues \cite{AckermanLeiserEtAl_2013_Is_comprehension_of_problem_solutions_resistant_to_misleading_heuristic_cues}. %Humans may resort to \textit{`fast reading'} because \textit{`slow reading'} is slow and mentally taxing, but AI systems should not have that limitation. % \cite{}. 

The previous generations of NLP models have already achieved high performance on many RC benchmarks, but they were found to often `read fast', i.e. rely on shallow patterns \cite{ChenBoltonEtAl_2016_A_Thorough_Examination_of_the_CNNDaily_Mail_Reading_Comprehension_Task,DBLP:conf/emnlp/JiaL17,RychalskaBasajEtAl_2018_Does_it_care_what_you_asked_Understanding_Importance_of_Verbs_in_Deep_Learning_QA_System}. Fine-tuned Transformer-based models \cite{DBLP:conf/naacl/DevlinCLT19} still have similar shortcomings \cite[][inter alia]{DBLP:conf/aaai/SugawaraSIA20,DBLP:conf/aaai/RogersKDR20,DBLP:conf/emnlp/SenS20,KassnerSchutze_2020_Negated_and_Misprimed_Probes_for_Pretrained_Language_Models_Birds_Can_Talk_But_Cannot_Fly} in RC, as well as other tasks \cite{DBLP:conf/acl/McCoyPL19,JinJinEtAl_2020_Is_BERT_Really_Robust_Strong_Baseline_for_Natural_Language_Attack_on_Text_Classification_and_Entailment}. 

\begin{figure}[!t]
\footnotesize
\begin{mdframed}[backgroundcolor=black!8,rightline=false,leftline=false]
 \textbf{Context}: \GB{Leo Strauss} was a political philosopher and classicist.
 \GB{He} was \RB{born} in \RB{Germany} ~... Thoughts on Machiavelli is a book by \GB{Leo Strauss}... \\
 \textbf{Question}: \RB{Where} was the author of Thoughts of Machiavelli \RB{born}? \\
 \textbf{Answer}: Germany
\end{mdframed}

    \caption{A sample question from the \squad\ \cite{DBLP:conf/emnlp/RajpurkarZLL16} dataset. \GB{green} tokens are the words that a reader relying on coreference resolution would take into account, and \RB{red} tokens are the words that could be used to answer the question with entity type matching.} 
    \label{fig:sample-squad}
\end{figure}

Consider the example in \autoref{fig:sample-squad}. A human reader would ideally construct the coreference chain resolving the pronoun `he' to `Leo Strauss'. A possible heuristic-based solution is entity type matching \cite{DBLP:conf/emnlp/JiaL17}: a model could observe that a `where' question can only be answered by a `location' and among two such entities (`Germany' and `United States') the correct answer (`Germany') is closer to `born'. Such heuristic reasoners will not generalize to unseen examples. Thus a key challenge in building trustworthy and explainable RC systems is to make sure their decisions are based on valid reasoning steps. However, it is difficult to establish: (a) what that reasoning should be; and (b) whether a blackbox system adheres to it. 

The present study proposes a framework for the analysis of RC models that includes: (a) defining the expected reasoning; (b) analysing model performance using explainability techniques. In particular, we contribute a case study for RC questions involving coreference resolution and comparison: we define the expected `reasoning' for them  (\cref{sec:reasoning}) and use a combination of saliency-based and counterfactual explanations (\cref{sec:methodology}) to analyze RC systems based on \bert\ and \roberta\ encoders of various sizes (\cref{sec:results}). Overall, we find that the larger models are more likely to rely on the `right' information, but even they seem to learn specific lexical patterns rather than underlying linguistic phenomena.

\section{When do RC Model `Understand' A Text?}
\label{sec:reasoning}

\subsection{Understanding in Humans}

The phenomenon of `natural language understanding' is not yet sufficiently well defined even for human speakers, although it is pursued by at least three different fields: philosophy of mind \cite[e.g.][]{Grimm_2021_Understanding,Dellsen_2020_Beyond_Explanation_Understanding_as_Dependency_Modelling}, psychology \cite[e.g.][]{Christianson_2016_When_language_comprehension_goes_wrong_for_right_reasons_Good-enough_underspecified_or_shallow_language_processing,Zwaan_2016_Situation_models_mental_simulations_and_abstract_concepts_in_discourse_comprehension}, and pedagogy \cite[e.g.][]{Lander_2010_edges_of_understanding,DuffinSimpson_2000_Search_for_Understanding}. We cannot do this topic justice within the scope of this paper, but let us briefly outline the key premises about human understanding that we rely on in our work:

\begin{itemize*}
    \item Understanding is not truth-connected: it is ``a merely psychological state'' \cite{Grimm_2012_The_Value_of_Understanding};
    \item Its objects are something like `connections' or `relations' of the phenomenon X to other phenomena \cite{Grimm_2021_Understanding};
    \item It is not binary: teachers routinely talk of `levels of understanding', `continuum of understanding' or `partial understanding' \cite{NurhudaRusdianaEtAl_2017_Analyzing_Students_Level_of_Understanding_on_Kinetic_Theory_of_Gases};
    \item It is different from `knowledge', i.a. since it is ``not transmissible\footnote{This is why, as any teacher knows from practice, simply presenting the students with definitions or principles does not necessarily result in understanding of those principles or definitions.} in the same sense as knowledge is'' \cite{BurnyeatBarnes_1980_Socrates_and_Jury_Paradoxes_in_Platos_Distinction_Between_Knowledge_and_True_Belief}. 
  %(Burnyeat 1980: 186)
 %the objects of understanding seem more structured and interconnected than objects of knowledge
\end{itemize*}

If human understanding is about establishing connections between new and existing conceptualizations, its success depends on the pre-existence of a suitable set of conceptualizations, to which the connections can be established (this is why e.g. algebra is taught in schools before differential calculus). The set of conceptualizations that each of us possesses is unique, since it depends on our experience of the world (cf. Fillmore's `semantics of understanding' \cite{Fillmore_1985_Frames_and_semantics_of_understanding}). This, together with other factors like level of motivation, attention etc., explains the variation in human understanding: we may grasp different sets of possible connections between different aspects of the new phenomenon and our pre-existing worldview.

\subsection{`Understanding' in Machines}

Much research on human understanding focuses on mechanisms that fundamentally do not apply to current NLP systems, such as the distinction between `knowledge' and `understanding' or the fact that humans will fail to understand if they don't have suitable pre-existing conceptualizations (while an encoder will encode text even if its weights are random). Since the mechanism (and its results) is so fundamentally different, terms like `natural language understanding' or `reading comprehension'\footnote{\citet{MarcusDavis_2019_Rebooting_AI_Building_Artificial_Intelligence_We_Can_Trust} dispute even the applicability of the term ``reading'', declaring the current QA/RC systems ``functionally illiterate'' since they cannot draw the implicit inferences crucial for human reading.} for the current NLP systems are arguably misleading. It would be more accurate to talk instead of `natural language processing' and `information retrieval'.

While terms like `understanding' are widely (mis)applied to models in AI research \cite{10.1145/3449639.3465421}, their definitions are scarce. Turing famously posited that the question ``can machines think?'' is too ill-defined to deserve serious consideration, and replaced it with a behavioral test (conversation with a human judge) for when we would say that thinking occurs \cite{Turing_1950_Computing_Machinery_and_Intelligence}. Conceptually, this is still the idea underlying the `NLU' benchmarks used today: we assume that for models to perform well on collections of tests such as GLUE \cite{WangSinghEtAl_2018_GLUE_Multi-Task_Benchmark_and_Analysis_Platform_for_Natural_Language_Understanding,WangPruksachatkunEtAl_2019_SuperGLUE_Stickier_Benchmark_for_General-Purpose_Language_Understanding_Systems}, some capacity for language understanding is required, and hence if our systems get increasingly higher scores on such behavioral tests, this would mean progress on `NLU'. However, just like the Turing test itself turned out to be ``highly gameable'' \cite{MarcusRossiEtAl_2016_Beyond_Turing_Test}, so are our tests\footnote{In fact, the larger the dataset, the more of likely spurious patterns are to occur \cite{GardnerMerrillEtAl_2021_Competency_Problems_On_Finding_and_Removing_Artifacts_in_Language_Data}. This presents a fundamental problem for data-hungry deep learning systems: ``the models, unable to discern the intentions of the data set’s designers, happily recapitulate any statistical patterns they find in the training data''
 \cite{Linzen_2020_How_Can_We_Accelerate_Progress_Towards_Human-like_Linguistic_Generalization}.} \cite[][inter alia]{DBLP:conf/aaai/SugawaraSIA20,DBLP:conf/aaai/RogersKDR20,DBLP:conf/emnlp/SenS20,KassnerSchutze_2020_Negated_and_Misprimed_Probes_for_Pretrained_Language_Models_Birds_Can_Talk_But_Cannot_Fly,DBLP:conf/acl/McCoyPL19,JinJinEtAl_2020_Is_BERT_Really_Robust_Strong_Baseline_for_Natural_Language_Attack_on_Text_Classification_and_Entailment}.

All this suggests that, at the very least, we need a better specification for the success criteria for such behavioral tests. Instead of asking \textit{``Does my RC model ``understand'' language?''} we could ask: \textit{``Does my RC model produce its output based on valid information retrieval and inference strategies?''} Then the next question is to specify what strategies would be valid and acceptable, which is possible to do on case-by-case basis.

With respect to `machine reading comprehension', a recent proposal by \citet{DBLP:conf/acl/DunietzBBRCF20} is based on whether a model can extract certain information that should be salient for a human reader (e.g. spatial, temporal, causal relations in a story). However, a model can extract such `right' information through a `wrong' process, e.g. some shallow heuristic. Hence the definition of `NLU' needs at least two components: (a) the specific information that the model is expected to be `extract' from the text; and (b) a valid process with which such `extraction' is performed. And this would still not be enough: the model could have simply memorized \textit{both} the right answer and the strategy to find it for some limited set of examples. We argue that the third key prerequisite is the ability to generalize: to \textit{consistently} use the `right' information-seeking strategy in novel contexts.\footnote{This does not preclude errors (humans make them too).}

Thus we propose the following general success criteria for NLP systems:

\begin{definition}
\label{def-nlu}
A NLP system has human-level competence with respect to its task X iff:
\begin{enumerate}[itemsep=1pt,parsep=0pt,before={\parskip=0pt},label=(\alph*)]
    \item it is able to correctly perform the task X (identify the target information in QA, correctly classify texts, generate an appropriate translation etc.);
    \item it does so by relying predominantly on information that a competent human speaker would also find relevant\footnote{Note that this leaves room for NLP systems to rely on patterns humans may not be even aware of, as long as such patterns are valid. E.g. if a system learned to make health outcome predictions based on latent information about unknown drug interactions, that would be the discovery of new knowledge that the experts would then accept -- but not if its predictions were based on a spurious correlation with Marvel movie release dates.};
    \item it does so consistently under distribution shifts that do not pose challenges to competent human speakers.
\end{enumerate}
\end{definition}

\begin{table*}[ht]
        \centering
\footnotesize        
\begin{tabular}{p{0.011\textwidth} p{0.27\textwidth} p{0.16\textwidth} p{0.42\textwidth}}
\toprule 
& \textbf{Example} & \textbf{Step} & \textbf{Relevant Spans} \\
\midrule
\multirow{3}{*}[-12ex]{\rotatebox[origin=c]{90}{Comparison}} & \multirow{3}{0.27\textwidth}[-3.5ex]{
\hspace{-2pt}\textbf{Context}: Blind Shaft is a 2003 film about a pair of brutal con artists operating in the illegal coal mines of present day northern China. The Mask Of Fu Manchu is a 1932 pre-Code adventure film directed by Charles Brabin. \newline 
 \textbf{Question}: Which film came out earlier, Blind Shaft or The Mask Of Fu Manchu? \newline
 \textbf{Answer}: The Mask Of Fu Manchu 
} & Interpreting the question & \textit{came out} relation: <film, release date> \newline film entities: \textit{Blind Shaft}, \textit{The Mask Of Fu Manchu} \newline \textit{earlier:} date comparison \newline \textbf{target:} $min($release date\textsubscript{\textit{Blind Shaft}}, release date\textsubscript{\textit{The Mask of Fu Manchu}}$)$ \\ 
\cmidrule{3-4}
& & Identifying relevant information through referential equality
%\footnote{Referential equality can not always be established by exact string matching between the surface forms and the types of the entities \cite{DBLP:journals/tacl/LammPAACSC21}}
& 
  \textit{Blind Shaft}\textsubscript{\textbf{q}} := \textit{Blind Shaft}\textbf{\textsubscript{c}}\newline
  \textit{The Mask Of Fu Manchu}\textsubscript{\textbf{q}} := \textit{The Mask Of Fu Manchu}\textsubscript{\textbf{c}}.\newline
  \textit{came out}\textbf{\textsubscript{q}} := <date, film> construction\textbf{\textsubscript{c}} \newline 
  release dates: <\textit{Blind Shaft}, \textit{2003}>, <\textit{The Mask Of Fu Manchu}, \textit{1932}>
  \\
\cmidrule{3-4}
& & Value comparison
 & \textbf{solution:} \textit{earlier}\textsubscript{\textbf{q}} := \textit{min}\textsubscript{\textbf{c}} \newline $min(1932, 2003) = 1932$ \\
\midrule
\multirow{3}{*}[-6ex]{\rotatebox[origin=c]{90}{Coreference}} & \multirow{2}{0.27\textwidth}[-3ex]{\textbf{Context:} Barack Obama was the 44th president of the US. He was born in Hawaii. \newline
\textbf{Question:} Who was born in Hawaii? \newline
\textbf{Answer:} Barack Obama.}
& Interpreting the question & \textit{born} relation: <person, location> \newline  \textit{Hawaii}: location \newline \textbf{target:} \textit{born}: <\textit{Hawaii}, UNK>\\
\cmidrule{3-4}
& & Identifying relevant information through referential equality & \textit{Hawaii}\textsubscript{\textbf{q}} := \textit{Hawaii}\textsubscript{\textbf{c}} 
\newline \textit{born} relation: <\textit{he, Hawaii}> \\ 
\cmidrule{3-4}
& & Coref. resolution & <\textit{Barack Obama, he}> \newline \textbf{solution:} \textit{born} <\textit{Barack Obama, Hawaii}> \\
\bottomrule 
\end{tabular}

\caption{The basic reasoning steps for answering comparison and coreference questions.}
\label{tab:reasoning-steps}
\end{table*}

\subsection{Reasoning an RC Model \textit{should} Perform}

The second principle in our Def.~\autoref{def-nlu} is that the model should rely on the `right' information. While models can discover patterns unknown to humans, a competent human reader should at least find such patterns relevant post-factum.

What information-seeking strategy is needed depends on the type of question and the context. \citet{DBLP:journals/corr/abs-2107-12708} propose a classification of RC `skills' into five main groups: situation/world modeling, different types of inference/logical reasoning, the ability to combine information in multi-step reasoning, knowing what kind of information is needed and where to find it, and interpreting/manipulating linguistic input. A single question may require the competency of several types of `skills'.

This study contributes an empirical investigation on two RC `skills' in the broad category of `interpreting/manipulating linguistic input': coreference resolution and comparison. 
Both of them rely on the contextual information and linguistic competence. Assuming that a human reader would first read the question and then read the context in order to find the answer, they would need to perform roughly three steps: (a) to interpret the `question' (akin to its transformation to a formal semantic representation or a query); (b) to identify the relevant information in the context through establishing the referential equality between expressions in the question and in the context; (c) to use that information to perform the operation of comparison or coreference resolution (see \autoref{tab:reasoning-steps}).\footnote{This definition could be developed further for more complex cases of coreference and comparison, or to model other variations of the human reading process, but this approximation suffices for our purposes and our RC data (see \cref{sec:dataset-models}).}

\subsection{Reasoning an RC Model \textit{does} Perform}
\label{sec:explainability}

Having established what reasoning steps an RC model \textit{should} perform, the next step would be to ascertain whether that is the case for specific models. But generally, the interpretability of DL models is an actively developed research area \cite{BelinkovGlass_2019_Analysis_Methods_in_Neural_Language_Processing_Survey,molnar2020interpretable}. In this study, we rely on a combination of two popular post-hoc explanation techniques, but we also discuss their limitations, and expect that new methods could soon be developed and used in the overall paradigm for the analysis of RC models that we propose.

\begin{figure}[!t]
  \centering
        \includegraphics[scale=0.7]{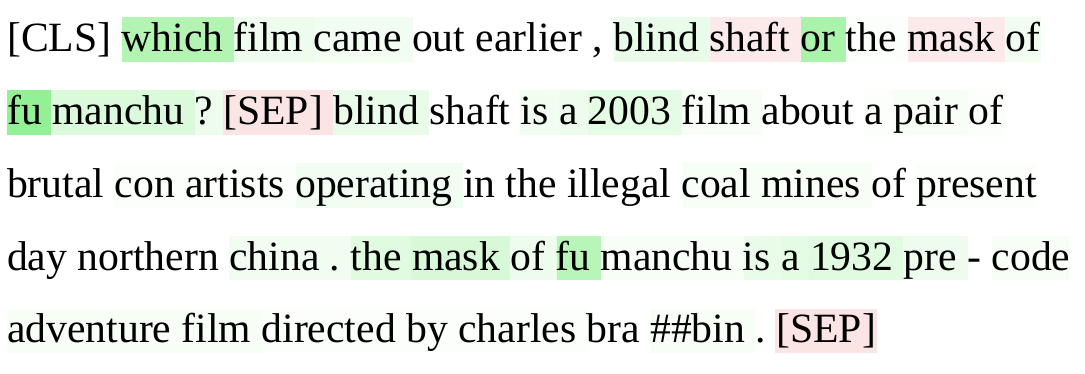}
    \caption{\ig\ saliency scores example. Green/red denotes positive/negative scores.}
    \label{fig:comparison-saliency}
\end{figure}

\textbf{Attribution/saliency-based methods} \citet{DBLP:journals/corr/LiMJ16a, DBLP:conf/icml/SundararajanTY17} provide a saliency score for each token in the input, which shows how `important' a given token is for the model decision in this instance. \autoref{fig:comparison-saliency} illustrates that such scores may not necessarily map onto human rationales.
% : is the token `which' important because it dictates the answer entity type or it is the first token?

To establish whether a model performs a given reasoning step (see \autoref{tab:reasoning-steps}), we define the following partition of the token space: the tokens the model \textit{should} find important (positive) vs the ones it \textit{should not} (negative). For example, to know if the model `attends' to the entities being compared, we can define the positive partition as \texttt{\string{blind, shaft, mask, of, fu, munchu\string}} and the negative partition as \texttt{\string{northern, china\string}}. If the model consistently follows this strategy, the average score should be higher for the positive rather than negative partition.

A limitation of saliency explanations is that they are not always faithful, i.e., do not reflect a model's true decision process \cite{DBLP:conf/emnlp/AtanasovaSLA20,Atanasova_Simonsen_Lioma_Augenstein_2022,DBLP:conf/emnlp/YeND21}. Also, even when they are faithful, i.e., when we can reliably say that a model places more `importance' on token $i$ than token $j$ in an instance, this does not imply that a \textit{set} of tokens $I$ is more salient than another set $J$. 

\textbf{Counterfactual explanations} have the form: ``had X not occurred, Y would not have occurred'' \cite{molnar2020interpretable}. In NLP, they are based on input perturbations \cite{DBLP:conf/iclr/KaushikHL20, gardner-etal-2020-evaluating, sen-etal-2021-counterfactually,10.1162/tacl_a_00486}. In our case, it translates to ``had the model not relied on information X, it could not have answered both the original and the perturbed instance correctly''. Thus the perturbation has to change the correct label, unlike for contrast sets \cite{gardner-etal-2020-evaluating}. 

Counterfactual (CF) explanations are considered to be more faithful, since they identify input features that impact predictions. However, they typically have to be manually generated \cite{DBLP:conf/iclr/KaushikHL20}, which makes large-scale \counterfactual\ generation prohibitively expensive \cite{DBLP:conf/emnlp/KhashabiKS20}.

We rely on both types of explanations as parallel sources of evidence about RC model reasoning, and define their alignment as follows:

\begin{definition}
\label{def-expl-alignment}
\textbf{Explanation Alignment.} A \counterfactual\ and saliency-based explanation \textit{align} when: (a) both the original and the counterfactually modified instance are answered correctly,\footnote{i.e. there is an exact match between the predicted and the correct answer.}; and (b) the positive partition has a statistical significantly higher average saliency score than the negative partition.  
\end{definition}

We define the alignment score as follows:

\begin{definition}
\label{def-alignment-score}
\textbf{Alignment Score}: The Alignment Score for a $<$dataset, model, reasoning step$>$ triple is the proportion of instances in that dataset for which different kinds of explanations align (according to our Def. \ref{def-expl-alignment}). 
\end{definition}

We interpret a high alignment score as evidence that both kinds of explanations are faithful, and the model indeed performs the expected reasoning steps.

\section{Methodology} 
\label{sec:methodology}

\subsection{Datasets and Models}
\label{sec:dataset-models}

For \textbf{Coreference}, we use the \quoref\ ~\cite{DBLP:conf/emnlp/DasigiLMSG19} dataset (20K training and 2.4K validation instances) where the annotators were asked to \textit{design} questions for a given text so that answering those would require resolving anaphora. For \textbf{Comparison}, we sample questions from \hotpot\ \cite{DBLP:conf/emnlp/Yang0ZBCSM18} and \wiki\ \cite{DBLP:conf/coling/HoNSA20}: two datasets with questions manually annotated with their reasoning type (bridge or comparison). We select the `comparison' questions containing comparative adjectives or adverbs in them (23K training, 3K validation instances). These resources are based on Wikipedia and have multiple passages as contexts, but the sentences (typically 2-3) necessary to answer a question are marked as `supporting facts'. Since we are not focusing on the multi-hop information retrieval skill, we limit the contexts to these sentences. 

We experiment with five pre-trained Transformer-based encoders of the BERT family: \rl~\cite{DBLP:journals/corr/abs-1907-11692}, \bertlc, \bertbc ~\cite{DBLP:conf/naacl/DevlinCLT19}, \bertmedium, and \bertsmall~\cite{DBLP:journals/corr/abs-1908-08962,bhargava2021generalization}. These \bert\ models differ mainly in the structure of architecture blocks and the number of parameters, while \roberta\ also has a different training corpus and optimization. Since larger models were shown to generalize better for some use cases \cite{hendrycks-etal-2020-pretrained, bhargava2021generalization}, we investigate whether they also are more likely to be right for the right reasons.

We \finetune\ each encoder using the architecture in \citet{DBLP:conf/naacl/DevlinCLT19} (see the appendix for details) and evaluate them on the validation set (as the test sets are not public). We use the standard evaluation metrics in extractive QA: \textbf{\fscore} (the percentage of token overlap between predicted and `gold' answers, averaged over all data points), and \textbf{\emscore} (the number of data points where the predicted answer matches the `gold' answer). 

\begin{figure*}[!t]
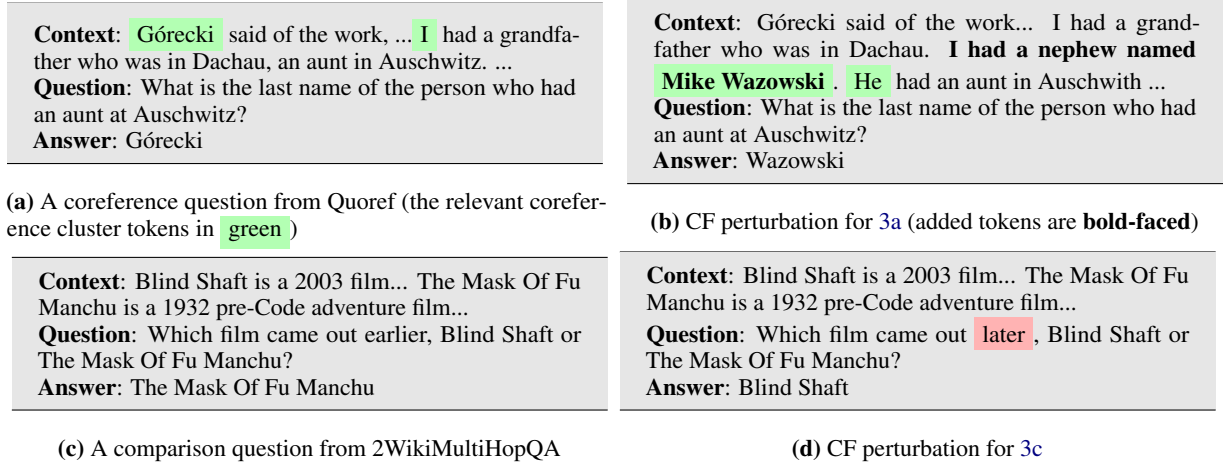

    \centering
    \begin{subfigure}[b]{0.49\textwidth}
        \centering
        \begin{mdframed}[backgroundcolor=black!10,rightline=false,leftline=false]
        \footnotesize
        \textbf{Context}: \GB{Górecki} said of the work, ...\GB{I} had a grandfather who was in Dachau, an aunt in Auschwitz. ...\\
        \textbf{Question}: What is the last name of the person who had an aunt at Auschwitz? \\
        \textbf{Answer}: Górecki
        \end{mdframed}
        \caption{A coreference question from \quoref\ (the relevant coreference cluster tokens in \GB{green})}
         \label{fig:coref-original}
    \end{subfigure}
    \hfill
    \begin{subfigure}[b]{0.49\textwidth}
        \centering

        \begin{mdframed}[backgroundcolor=black!10,rightline=false,leftline=false]
        \footnotesize
        \textbf{Context}: Górecki said of the work... I had a grandfather who was in Dachau. \textbf{I had a nephew named \GB{Mike Wazowski}}. \GB{He} had an aunt in Auschwith ...\\ 
        \textbf{Question}: What is the last name of the person who had an aunt at Auschwitz? \\
        \textbf{Answer}: Wazowski
        \end{mdframed}
        \caption{\counterfactual\ perturbation for \ref{fig:coref-original} (added tokens are \textbf{bold-faced})}
        \label{fig:coref-perturbed}
    \vspace{.4em}
    \end{subfigure}
    
    \hfill
    \begin{subfigure}[b]{0.49\textwidth}
        \centering
        \begin{mdframed}[backgroundcolor=black!10,rightline=false,leftline=false]
        \footnotesize
        \textbf{Context}: Blind Shaft is a 2003 film...
        The Mask Of Fu Manchu is a 1932 pre-Code adventure film... \\
        \textbf{Question}: Which film came out earlier, Blind Shaft or The Mask Of Fu Manchu? \\
        \textbf{Answer}: The Mask Of Fu Manchu 
        \end{mdframed}
        \caption{A comparison question from \wiki\ }
         \label{fig:comparison-original}
    \end{subfigure}
    \hfill    
    \begin{subfigure}[b]{0.49\textwidth}
        \centering
        \begin{mdframed}[backgroundcolor=black!10,rightline=false,leftline=false]
        \footnotesize
        \textbf{Context}: Blind Shaft is a 2003 film...
        The Mask Of Fu Manchu is a 1932 pre-Code adventure film... \\
        \textbf{Question}: Which film came out \RB{later}, Blind Shaft or The Mask Of Fu Manchu?\\
        \textbf{Answer}: Blind Shaft 
        \end{mdframed}
        \caption{\counterfactual\ perturbation for \ref{fig:comparison-original}}
         \label{fig:comparison-cf-comparative-adjective}
    \end{subfigure}
\label{fig:counterfactual-perturbations}
\caption{Examples of \counterfactual\ perturbations used in this study.}
\end{figure*}

\subsection{Counterfactual Explanations}
\label{sec:counterfactual-exp}

Our formulation of reasoning (\autoref{tab:reasoning-steps}) consists of three basic steps for both coreference and comparison: interpreting the question, identifying the relevant information through referential equality, and the target operation on the identified information (coreference resolution or value comparison). We focus on the final step, since: (a) it implicitly relies on correct semantic parsing of the question and the context; (b) referential equality in our data is in large part trivial: most entities have the same surface form in the question and the text.

An obvious semantically valid perturbation that should change the prediction (and thus test for the model's understanding of the comparison operation) is to replace the comparative adjectives with their antonyms (\autoref{fig:comparison-cf-comparative-adjective}). Since our sample only contains 6 tokens used as comparison operators, we define appropriate replacements manually.\footnote{earlier$\leftrightarrow$later, first$\rightarrow$later, more recently$\rightarrow$earlier, older$\leftrightarrow$younger.}

For coreference questions, a competent RC model would at least resolve the coreference chain for the target entity. A context can have many coreference clusters, so we need to identify the relevant one. In the \quoref\ dataset, we use the instances where the relevant cluster itself contains the answer entity\footnote{We extract the clusters using an off-the-shelf coreference resolver \cite{DBLP:conf/emnlp/ClarkM16} implemented in \href{https://spacy.io}{Spacy}.} (see \autoref{fig:coref-original}), and therefore, can be extracted automatically. This leaves us with $55\%  (1329/2418)$ of the validation instances. These are further subsampled to manually create 100 \counterfactual\ instances by inserting a new sentence, which includes the new and excludes the old answer (see \autoref{fig:coref-perturbed}). Similarly to the comparison questions, the original answer entity remains in the context. If the model uses the `shortcut' of choosing the most frequent entity in the context \cite{DBLP:conf/acl/WuMRG20}, it should not be able to answer both the original and the perturbed instance correctly.

\begin{figure}[!t]
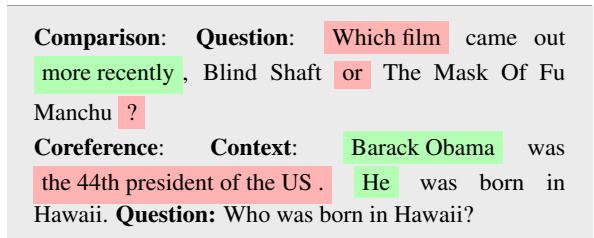

\footnotesize
\begin{mdframed}[backgroundcolor=black!8,rightline=false,leftline=false]
 \textbf{Comparison}: \textbf{Question}: \RB{Which film} came out \GB{more recently}, Blind Shaft \RB{or} The Mask Of Fu Manchu \RB{?} \\
 \hrulefill
 \textbf{Coreference}: \textbf{Context}: \GB{Barack Obama} was \RB{the 44th president of the US .} \GB{He} was born in Hawaii. \textbf{Question:} Who was born in Hawaii?
\end{mdframed}
    \caption{\GB{Positive} and \RB{negative} partitions for saliency explanations.} 
    \label{fig:token-partitions}
\end{figure}

\subsection{Saliency-based Explanations}
\label{sec:saliency-exp}

We obtain token saliency scores from two families of attribution/saliency methods: \occlusion\ \cite{DBLP:conf/acl/DeYoungJRLXSW20}, a method based on perturbations, and Integrated Gradients (\ig,  \citet{DBLP:conf/icml/SundararajanTY17}), a method based on gradients.\footnote{ \citet{DBLP:conf/emnlp/AtanasovaSLA20} shows that for Transformer based architectures, \occlusion\ is the best perturbation method by two evaluation criteria: agreement with human rationale and faithfulness. A recent paper by \citet{DBLP:conf/emnlp/YeND21} finds \ig\ to be one of the most faithful gradient-based methods for extractive QA, only outperformed by Layerwise Attention Attribution (LAA), a method proposed in the paper itself. We leave LAA and other popular explainability methods such as LIME \cite{DBLP:conf/naacl/Ribeiro0G16} for future work.}

\textbf{Design decisions:} RC models typically predict two scores ($t_s, t_e$) for each token $t$: the probability of $t$ being the start and the end of the answer span. Any attribution method produces two scores ($A_{start}^t$, $A_{end}^t$) for each token $t$, indicating how `important' $t$ is for predicting the start/end of the answer span. Following \citet{DBLP:journals/corr/abs-2009-07896}, we use $A_{start}$ in all our saliency experiments.\footnote{We also briefly experimented with ($A_{start} + A_{end}$)/2 for \occlusion\, but it yielded very similar saliency ranking of the tokens on a 100 sample subset of the Comparison dataset.}

For \occlusion, we calculate $A_{start}^t$ by replacing $t$ in the input with a baseline token (\texttt{MASK}) and measuring the change in $t_s$. DNNs map an input vector to a scalar value (loss/ class probability). Gradient-based methods measure $A_{start}^t$ using the gradient of the token $t$ w.r.t. this scalar function (we use $argmax(t_s)$). \ig\ sums these gradient values along a linear path from a baseline to the current instance. Both \occlusion\ and \ig\ need a baseline token, which for us is the \texttt{MASK} token.

Gradient-based methods in NLP do not produce a scalar saliency score, i.e., $A_{start}^t$ is a vector because the input is an embedding \textit{matrix} and not a vector. Two common ways to summarize this vector to a scalar are: (a) scalar product between the input and the gradient vector \cite{DBLP:conf/acl/HanWT20}; or (b) $l_p$ norm, where $p \in {1, 2}$ \cite{DBLP:conf/emnlp/AtanasovaSLA20}. We use $l_2$ norm (see the discussion in \cref{sec:alignment}).

\textbf{Token partitions:} \autoref{fig:token-partitions} shows the token partitions used for the same reasoning steps (comparison and coreference resolution) that we also target with the \counterfactual\ perturbations. For comparison the positive partition consists of the \textbf{question} token(s) expressing the comparison operation (e.g. `more recently'). The negative partition consists of question tokens that are not in the set of entities or values that need to be compared, or in the set of verbs (which could capture the relation between the entities and their values). For coreference resolution, the positive partition is the \textbf{context} tokens in the relevant coreference cluster (\cref{sec:counterfactual-exp}). The negative partition is the set of context tokens that are not in: (a) the positive partition; and (b) match the question tokens.

\section{Results \& Analysis}
\label{sec:results}

\subsection{Base Model Performance}
\label{sec:base-model-perf}

As a sanity check, we fine-tune all models on the data described in \cref{sec:dataset-models} (\autoref{tab:base-results}). 
For coreference, the \fscore\ of our best model (\rl) is slightly better (82.10) than the previously reported score (79.64, \citet{DBLP:conf/acl/WuMRG20}). The comparison instances are sampled from \textit{parts} of two datasets, and so a direct comparison is not possible.\footnote{The best model (\rl) has an \fscore\ of 92\%, slightly better than the highest score reported on the \href{https://hotpotqa.github.io/}{\hotpot\ leaderboard} (89.14\%) and much better than the baseline model for the \wiki\ dataset (65.02, \cite{DBLP:conf/coling/HoNSA20}).} 

The size and the model family matter: \roberta\ performs better than \bert\ for two models of the same size, and the larger models do better. Interestingly, the difference is more pronounced for the \quoref\ dataset, where the instances have longer contexts and the questions are more complex.

\begin{table}[!t]
\footnotesize
\centering
\begin{tabular}{@{}lllll@{}}
\toprule

                                           & \multicolumn{2}{l}{Comparison}                                                   & \multicolumn{2}{l}{Coreference}                                  \\ \midrule
                                           & F1 & EM & F1 & EM \\ \midrule
\rl\         & \GB{92.08}                                  & \GB{91.07}                                   & \GB{82.10}                                  & \GB{79.39}                   \\
\bertlc\     & 89.23                                  & 88.57                                   & 71.91                                  & 68.47                   \\
\bertbc\     & 89.38                                  & 88.37                                   & 64.62                                  & 59.38                   \\
\bertmedium\ & 86.45                                  & 85.96                                   & 60.16                                  & 54.82                   \\
\bertsmall\  & 71.44                                  & 69.87                                   & 50.94                                  & 43.39                   \\ \bottomrule
\end{tabular}
\caption{Average (3 runs) results %(F1=\fscore\, EM=\emscore, as defined before) 
of different models on Comparison and Coreference datasets. The STD varies between $0.01-0.72\%$. \GB{Green} indicates the best scores.}
\label{tab:base-results}
\end{table}

\subsection{Counterfactual Explanations}
\label{sec:cf-results}
\autoref{tab:comp-counterfactual} compares the \fscore\ of the original (`og') vs counterfactual (`cf') instances. For the comparison questions, the performance on the original and \counterfactual\ instances are very close for all models except \bertsmall. Bigger models consistently perform better, but in most cases the difference with the next larger model is relatively small.

For coreference questions, \counterfactual\ instances are much more difficult for all models. Even the best model \rl\ experiences a $24\%$ drop. All \bert\ models perform poorly: even the larger ones have a $40\%$ performance drop (\bertlc). Thus, the \counterfactual\ tests show that the models are more likely to follow the expected reasoning strategy for comparison, but not for the coreference questions.

\begin{table}[!t]
\footnotesize
\centering
\begin{tabular}{llllll}
\toprule
  & \multicolumn{2}{l}{Coreference} & \multicolumn{2}{l}{Comparison} \\ \midrule
  & og    & cf                & og                  & cf                              \\
\rl\ & 92.0  & \RB{70.7}  & 99.4                & 98.9                            \\
\bertlc\ & 86.2  & \RB{50.8}  & 98.9                & 93.1                            \\
\bertbc\ & 82.5  & \RB{39.2}  & 98.4                & 91.8                            \\
\bertmedium\ & 74.0  & \RB{35.8}  & 97.4                & 96.5                            \\
\bertsmall\ & 67.2  & \RB{29.4}  & 68.2                & \RB{45.3}               \\ \bottomrule
\end{tabular}
\caption{\fscore\ for the original and the \counterfactual\ perturbations. \RB{Red} denotes significant drop.}
\label{tab:comp-counterfactual}
\end{table}

\subsection{Alignment Score} 
\label{sec:alignment}

For statistical significance testing in `Expectation Alignment' (Def. \ref{def-expl-alignment}), we use a one-tailed independent t\_test ($p=0.05$) with the null hypothesis that \textit{the positive partition does not have a higher average saliency score}. \autoref{tab:saliency-alignment} shows the `Alignment Score' (Def. \ref{def-alignment-score}) results for comparison and coreference resolution (\cref{sec:saliency-exp}), using saliency scores from \ig\ and \occlusion. 

Ideally, for a random partition of tokens in any instance, the positive and the negative partitions should have similar saliency scores. For a dataset, they should be \textit{significantly} different in $\approx$ $0\%$ cases.\footnote{Aggregation of local explanations such as saliency scores are not \textit{guaranteed} to produce faithful global explanations \cite{DBLP:journals/ai/SetzuGMTPG21}, but this is a convincing evidence.} For \occlusion, the saliency scores are significantly different in only $5.6-8.2\%$ instances for a random partition. Recall that in \cref{sec:saliency-exp} we discussed 3 summarizers for \ig. Among all of them, the $l_2$ norm is the only one where this happens in $5.2-7.3\%$ cases, for the other two the numbers are between $11.3-28.9\%$.

\begin{table}[!t]
\footnotesize
\centering
\begin{tabular}{p{0.35\linewidth} p{0.1\linewidth} p{0.1\linewidth} p{0.1\linewidth} p{0.1\linewidth}}
\toprule
& \multicolumn{2}{p{0.2\linewidth}}{Coreference} & \multicolumn{2}{p{0.2\linewidth}}{Comparison}\\ 
\midrule
      & IG               & Occ                 & IG               & Occ \\ \midrule
\rl\    & 33.3             & 69.7                & 33.8             & 67.0 \\
\bertlc\    & 12.5             & 58.3                & 34.1             & 65.9\\
\bertbc\   & 21.4             & 42.9           & \GB{83.8}        & 69.0\\
\bertmedium\    & \GB{81.8}        & 36.4                & \GB{82.2}        & 42.0\\
\bertsmall\    & \GB{83.3}        & 33.3                & \GB{86.3}        & 16.3\\ \bottomrule
\end{tabular}
\caption{Alignment score between counterfactual explanations vs \ig\ (Integrated Gradients) or Occ (Occlusion). \GB{Green} indicates methods with $>80\%$ alignment.}
\label{tab:saliency-alignment}
\end{table}

\autoref{tab:saliency-alignment} shows that, counter-intuitively, for both comparison and coreference questions the larger models overall have \textit{lower} \ig\ alignment scores, meaning that they do not pay as much `attention' to the tokens we defined as important. This is despite the fact that for comparison the above \counterfactual\ experiment suggests that the models do perform the expected reasoning operations. One possible explanation is that \ig\ simply does not reliably capture the model's reasoning process, and \occlusion\ does better at that because its trend in alignment is the opposite of \ig: bigger models tend to have significantly higher alignment scores.\footnote{The lack of alignment between the two techniques is consistent with the findings of \citet{DBLP:conf/emnlp/AtanasovaSLA20}.}

Another possible explanation is that \ig\ explanations are in fact faithful, but, having more `attention' to the tokens we defined as important is counter-productive. Consider that the \bertsmall\ model achieves an \emscore\ of $87\%$ on the original questions containing the comparative tokens `earlier', `first' and `older' (which are 2.1 times more frequent in the training data than all others), and an \emscore\ of $28\%$ on the other original questions. Yet overall the model performs poorly, and thus the reliance on these highly frequent comparative adjectives could be a bug rather than a feature. As this hypothesis brings into question the overall utility of saliency-based explanations for testing for the `correct' reasoning steps, we hope it will be investigated in more depth in future work.

\subsection{Generalization Tests}
\label{sec:generalization-tests}

\autoref{tab:comp-counterfactual} shows that when measured with CF tests, most models do not follow the expected coreference resolution strategy, but they do so for comparison. Still, based on our success criteria (Def. \autoref{def-nlu}), we cannot yet conclude that they `understand' comparison. A human would be able to disassociate the logical operation of comparison from the surface realizations, i.e., they would be able to answer a question correctly with either of the surface forms `younger' and `more junior'.

For the \counterfactual\ experiments reported up until this point the perturbations were in-distribution, i.e., the training data had both the original question ``who is younger'' and the \counterfactual\ ``who is older''. Now we replace the comparative adjectives with antonyms that are not in the training data (see the appendix for details). We also increase the context size by using full paragraphs instead of just the sentences marked as `supporting facts', to see if the models would be `distracted' by more information.

\autoref{tab:comp-counterfactual-gen} shows a considerable drop in performance for CF-ood condition for all models. The larger models generalize better: \rl\ and \bertlc\ perform $2\%$ and $8\%$ worse for CF questions, whereas \bertsmall\ exhibits a $29\%$ reduction. The `supporting facts only' condition is overall easier than the `paragraphs' condition.

\begin{table}[!t]
\footnotesize
\centering
\begin{tabular}{p{0.25\linewidth} p{0.06\linewidth} p{0.06\linewidth} p{0.06\linewidth} p{0.06\linewidth} p{0.06\linewidth} p{0.05\linewidth}}
\toprule
& \multicolumn{3}{p{0.18\linewidth}}{Supporting Facts} & \multicolumn{3}{p{0.18\linewidth}}{Paragraphs}\\ 
\midrule
  & OG      & CF            & CF-ood                         & OG                   & CF            & CF-ood                          \\ \midrule
\rl\ & 99.4    & 98.9           & \RB{77.2}         & 98.7                 & 96.4        & \RB{74.8}          \\
\bertlc\ & 98.9    & 93.1            & \RB{68.7}         & 98.0                 & 90.8        & \RB{67.5}           \\
\bertbc\ & 98.4    & 91.8            & \RB{58.1}         & 97.0                 & \RB{86.8}      & \RB{59.9}           \\
\bertmedium\ & 97.4    & 96.5            & \RB{64.4}         & 96.2                 & \RB{86.3}      & \RB{66.3}           \\
\bertsmall\ & 68.2    & \RB{45.3}            & \RB{57.1}         & 68.3                 & \RB{47.6}      & \RB{58.8}           \\ \bottomrule
\end{tabular}
\caption{\fscore\ for the original (OG) comparison questions and their counterfactual perturbations in (CF) and out (CF-ood) of the training distribution. The models are provided either a smaller context of supporting facts or full paragraphs. %`CF-ood' denotes \counterfactual\ perturbations with out-of-distribution synonyms. 
\RB{Red} indicates a significant drop in performance.}
\label{tab:comp-counterfactual-gen}
\end{table}

\subsection{Heuristics for Coreference Questions}
\label{sec:heuristics}

Since the \counterfactual\ tests (\cref{sec:cf-results}) do not show that \bert\ models can cope with the altered coreference chains, we have to conclude that they do not follow the expected reasoning steps. Though given the above-chance performance they must follow some other strategy. We test the hypothesis that many of the coreference questions can be answered by simple heuristics and that the models resort to those. Specifically, we define an unsupervised dataset-independent heuristic method consisting of two steps: sentence selection and phrase extraction.

\textbf{Sentence Selection}: Among all the context sentences $\{c_i\}$, select the one that is the `closest' to the question $q$. We experiment with 4 options for similarity: \textbf{token-overlap} (number of common tokens in $q$ and $c_i$), \textbf{sentence encoder} (cosine similarity between the sentence embeddings of $q$ and $c_i$ created by a sentence encoder \cite{reimers-gurevych-2019-sentence}), \textbf{LCS} (number of tokens in the Longest Common Subsequence between $q$ and $c_i$), and \textbf{position} (simply taking the first sentence in the context following \citet{DBLP:conf/emnlp/KoLKKK20}).

\begin{table}[!t]
\footnotesize
\centering
\begin{tabular}{lllll}
\toprule
                                                           & \multicolumn{2}{l}{Coreference} & \multicolumn{2}{l}{\squad}       \\ \midrule
                                                           & \fscore             & EM             & \fscore             & EM            \\ \midrule
\begin{tabular}[c]{@{}l@{}}Token\\ overlap\end{tabular}    & \GB{21.5}  & \GB{12.9}  & \GB{26.68} & \GB{21.64} \\
LCS                                                        & 17.2           & 12.9           & 19.59          & 15.97          \\
Position                                                        & 12.3           & 7.9           & 21.62          & 16.32          \\
\begin{tabular}[c]{@{}l@{}}Sentence\\ encoder\end{tabular} & 20.43          & 9.67           & 25.91          & 20.61          \\ \bottomrule
\end{tabular}
\caption{Results for different heuristic methods on the coreference and \squad\ datasets. \GB{Green} indicates the best score.}
\label{tab:heuristic}
\end{table}

\textbf{Phrase Extraction}: We assume that the model would also learn to look for a named entity in the selected sentence. The question dictates the \textit{type} of this entity (e.g. `where' $\rightarrow$ location, `who' $\rightarrow$ person name). The type could be determined by a simple mapping between `wh' question words and entity types, but this can fail (e.g. for the question ``Who won the World Cup in 2002?'' the expected answer is a location, not a person). Therefore, we fine-tune a Transformer model to predict the answer type from the question.\footnote{The accuracy for this model is $85.7\%$. See the appendix for results from multiple models and loss functions.}

\autoref{tab:heuristic} shows the best heuristic has an \fscore\ of $21.5\%$ on the coreference dataset, and $26.68\%$ on \squad ~\cite{DBLP:conf/emnlp/RajpurkarZLL16}, which we use for validation. The \squad\ score is comparable to the previously reported result of $26.7\%$ in \citet{DBLP:conf/emnlp/SenS20} for an algorithm predicting entity types heuristically, and choosing the entity from the whole context instead of the best possible sentence. \citet{choudhury2022epqa} uses a similar approach to find \quoref\ questions that can be answered heuristically, but our algorithm has more sentence selection strategies, and unlike ours, \citet{choudhury2022epqa} only uses one loss function in the phrase extraction model.

Nevertheless, the best heuristic algorithm performs considerably worse than the smallest \bertsmall\ model ($51\%$, \autoref{tab:base-results}). Performance alone cannot reveal whether this strategy is used in the instances where it \textit{would} be sufficient, but this result shows that even the smaller models must either rely on a more successful (but still imperfect) strategy, or at least rely on more than one heuristic. The problem with discovering potential `shortcuts' in low-performing models is complicated as these strategies are not necessarily human-interpretable: \citet{GonzalezRogersEtAl_2021_On_Interaction_of_Belief_Bias_and_Explanations} show that humans struggle to predict the answer chosen by poorly performing RC models, even when the saliency explanations for that answer is shown, because these answers simply do not align with human RC strategies. 

\section{Discussion and Related Work}
\label{sec:related} 

Our work continues the emerging trend of research on being `right for the right reasons'  \cite[][inter alia]{DBLP:conf/acl/McCoyPL19,DBLP:conf/naacl/ChenD19,DBLP:conf/acl/MinWSGHZ19,10.1162/tacl_a_00486}. We contribute stricter success criteria for behavioral tests of NLP models (Def. \autoref{def-nlu}), and, for the RC task, develop the methodology of: (a) defining what information the model should rely on for a given linguistic, logical, or world knowledge `skill'; (b) systematically testing the behavior of RC models with interpretability techniques for whether they rely on that information. This is most closely related to the work on `defining comprehension' by \citet{DBLP:conf/acl/DunietzBBRCF20}, though their testing is limited to probing the models with RC questions. Another related study is the QED framework \cite{DBLP:journals/tacl/LammPAACSC21}, annotating Natural Questions \cite{KwiatkowskiPalomakiEtAl_2019_Natural_Questions_Benchmark_for_Question_Answering_Research} with `explanations' of the expected reasoning process. Their expected reasoning process also contains 3 steps, partly similar to ours: selecting a relevant sentence, referential equality, and deciding on whether this sentence entails the predicate in the question. However, the goals of QED are to: (a) predict both the answer and the explanation for a question; and (b) understand if explanations help QA models. Such explanation annotations are unavailable for most datasets, and few QA models produce explanations. Therefore, our approach of: (a) defining expected reasoning steps; and (b) using model interpretations to validate such steps applies to a broader class of models.

This study is also related to the overall efforts to define what kinds of `skills' RC models can be expected to exhibit \cite{DBLP:conf/emnlp/SugawaraISA18,DBLP:conf/lrec/SchlegelVFNB20,DBLP:journals/corr/abs-2107-12708}. While these works focus on the high-level taxonomies of `skills', we contribute practical definitions for two linguistic `skills' (comparison and coreference resolution) which could be used for analyzing model performance. Implicitly, research proposing RC resources that target various specific `skills' (e.g. TempQuestions \cite{JiaAbujabalEtAl_2018_TempQuestions_Benchmark_for_Temporal_Question_Answering} for temporal order, MathQA \cite{AminiGabrielEtAl_2019_MathQA_Towards_Interpretable_Math_Word_Problem_Solving_with_Operation-Based_Formalisms} for numerical reasoning, etc.) also contributes to this area, but they typically rely on broad linguistic definitions rather than on steps for machine reasoning.

The saliency techniques we rely on have previously been used for extractive QA \cite{DBLP:journals/corr/abs-2110-08412}, but we are among the first \cite{DBLP:conf/emnlp/YeND21} to investigate their correlation with counterfactual explanations. For counterfactual perturbations, we also ensure that the perturbations are human-interpretable and change the prediction, which is not the case for adding incomprehensible text \cite{DBLP:conf/emnlp/KaushikL18}, removing words from questions, shuffling the context \cite{DBLP:conf/emnlp/SenS20}, or replacing context tokens with random tokens \cite{DBLP:conf/aaai/SugawaraSIA20}.

\section{Conclusion}

Making progress towards trustworthy NLP models requires specific definitions for the behavior expected of these models in different situations. We propose a framework for RC model analysis that involves: (a)~the definition of the expected `reasoning' steps; (b)~analysis of model behavior. We contribute such definitions for two linguistic `skills' (comparison and coreference resolution), and use parallel explainability techniques to investigate whether RC models based on BERT family encoders answer such questions correctly for the right reasons. We find that to be the case for comparison, but not for coreference. Moreover, we find that, even for comparison, the models `break' when encountering out-of-distribution counterfactual perturbations, suggesting that they memorize specific lexical patterns rather than learn more general reasoning `skills'. As such, more research is needed on developing definitions and tests for specific `skills' expected of NLU models, as well as on more faithful interpretability techniques.

\section{Acknowledgements}

We would like to thank Pepa Atanasova, Gary Marcus, Mark Steedman, and Bonnie Webber for the discussion of various aspects of this work. We also thank the anonymous reviewers for their time and insightful comments.

\begin{comment}
\section*{Impact Statement}

\paragraph{Broader impact.} The study focuses on methodology for determining when 

\paragraph{Personal data.} 

\paragraph{Potential risks.} 

\paragraph{Intended use.} The code to reproduce our results will be made publicly available for research purposes.

\paragraph{Institutional approval.} The study was approved by the Research Ethics Committee at the authors' institution.
\end{comment}

\bibliography{acl}
\bibliographystyle{acl_natbib}

\clearpage
\appendix

\section{Appendix}

\subsection{QA Model Training}
\label{sec:appendix-training}

For training the QA models in \cref{sec:dataset-models} The questions and contexts are concatenated, and a linear layer on top of the encoder is used to predict the probability of a context token $i$ being the start ($P_{i, s}$) or end ($P_{i, e}$) of an answer. The score ($S_{i,j}$) for a span with start token $i$ and end token $j$ is computed as $P_{i, s} + P_{j, e}$. For all valid combination of $i$ and $j$, the span with the highest score is chosen as the answer. A cross entropy loss between the actual and predicted start/end positions is minimized.

The models were trained for 10 epochs with a batch size of 16 using the Adam optimizer \cite{DBLP:journals/corr/KingmaB14} ($\beta_1=0.9, \beta_2=0.99, \epsilon=\texttt{1e-8}$, \texttt{weight\_decay = $0.01$}) and gradient clipping. The learning rate (LR) was kept at \texttt{1e-05} with a linear warm-up schedule (staring LR=$0$). The models were evaluated on a subset of the validation data every 500 mini-batches with early stopping on 100 evaluations \cite{pruksachatkun2020jiant}. The LR and batch size was determined by a small grid search on the coreference dataset: LR=\texttt{\{1e-05, 1e-04, 1e-03\}},  batch size = $\{8, 16, 32\}$. 

\subsection{Antonym Replacements for CF Generation}
\label{sec:appendix-antonym-replacements}

The antonym replacements for the generalization test (\cref{sec:generalization-tests}) are described below:
\begin{itemize}
\item first $\rightarrow$ less recently
\item older $\rightarrow$ less old, more junior, less mature, less grown-up
\item earlier $\rightarrow$ subsequently, thereafter, less recently
\item later $\rightarrow$ less recently
\item younger $\rightarrow$ more old, less junior, more mature, more grown-up
\item more recently $\rightarrow$ less recently, longer ago
\end{itemize}

\subsection{Supervised Entity Type Predictor}
\label{sec:appendix-supervised-entity-type}

Our goal is to build a classifier to predict the answer entity type from the question (\cref{sec:heuristics}). A sample data point is shown in \autoref{fig:sample-classification}. The entity types are defined in the Ontonotes-5 dataset \cite{pradhan-etal-2013-towards}. The answer entity type is detected from the context using an off-the-shelf entity detector implemented in Spacy.\footnote{\url{https://spacy.io}} When the answer is not a named entity, or the entity detector fails to determine its type, that question is discarded.

The classification models are trained on the \textbf{training} portion of \quoref\ and \squad\, which is further divided into train/dev/test (70/20/10) split for training and evaluation. The distribution of the class labels is very skewed.

\begin{figure}[!t]
\footnotesize
\begin{mdframed}[backgroundcolor=black!8,rightline=false,leftline=false]
\textbf{Text}: What is the full name of the person who is the television reporter that brings in a priest versed in Catholic exorcism rites? \\
\textbf{Label}: PER \\
\end{mdframed}
    \caption{A sample instance for answer entity type classifier.} 
    \label{fig:sample-classification}
\end{figure}

\textbf{Models}: We use two types of models: 1) a \finetuned\ 12 layer 768 dimensional \bertbc\ model; and 2) a popular word convolutional model for sentence classification \cite{kim-2014-convolutional} using three parallel filters (size 3, 4, and 5) and 300 dimensional Google News Word2Vec representations \cite{DBLP:journals/corr/abs-1301-3781}. 

\textbf{BERT model}: This model is trained for 5 epochs, with Adam optimizer \cite{DBLP:journals/corr/KingmaB14} with a weight decay of \texttt{1.0e-08} and a learning rate of \texttt{1.0e-05}. The sequence max length is kept at 128. We search for two hyperparameters: 1) number of epochs: 3-7, increasing by 1; and 2) learning rate: \texttt{1.0e-05}, \texttt{5.0e-05}, \texttt{1.0e-04}. 

\textbf{WordConv model}: This model is trained for 40 epochs, with Adadelta optimizer \cite{DBLP:journals/corr/abs-1212-5701} with a learning rate of \texttt{1.0e-05}. The sequence max length is again kept at 128. 

For both models, accuracy was used as the early stopping metric. We minimized the cross entropy (CE) loss in general, but for the WordConv model, a weighted CE loss was also implemented to account for the training data class-imbalance in \quoref. That did not improve the results significantly and was not used in the \bertbc\ model. \autoref{tab:supervised-etype-results} shows the detailed results. Finally, we choose the \finetuned\ \bertbc\ model as the entity detector as it performs the best. \citet{choudhury2022epqa} also proposes a model to determine the answer entity type from a question, but the major difference is the label space. The model in \citet{choudhury2022epqa} is trained to predict a label of ``UNKNOWN\_ENTITY" when the answer span is a) not a named entity or b) the entity detector can not find its type. However, an ``UNKNOWN\_ENTITY" label does not help the final algorithm (heuristic answer selection) to find the correct answer span. Therefore, our model never predicts this label, and consequently, has a better accuracy than \citet{choudhury2022epqa}. It potentially makes a mistake on the test data points that fall in the previous two categories, but the final algorithm is no worse than \citet{choudhury2022epqa}.

\begin{table}[!t]
\footnotesize
\centering
\begin{tabular}{lllll}
\toprule
Dataset & Model & Accuracy & \begin{tabular}[c]{@{}l@{}}Macro\\ F1\end{tabular} \\ \midrule
\multirow{2}{*}{\squad} & \bertbc\ & \GB{76.4} & \GB{56.2}\\
 & WordConv & 72.4 & 44.9\\ \midrule
\multirow{3}{*}{Coref} & \bertbc\ & \GB{85.7} & \GB{73.9}\\
 & WordConv & 85.0 & 67.6\\
 & \begin{tabular}[c]{@{}l@{}}WordConv\\ Weighted BCE\end{tabular} & 85.3 & 69.7\\ \bottomrule 
\end{tabular}%
\caption{Models for supervised entity type selection. \GB{Green} indicates the best results.}
\label{tab:supervised-etype-results}
\end{table}

\end{document}